\documentclass{article} %
\usepackage{iclr2027_conference,times}

\usepackage{amsmath,amsfonts,bm}

\def\figref#1{figure~\ref{#1}}

\def\secref#1{section~\ref{#1}}

\def\eqref#1{equation~\ref{#1}}

\def\algref#1{algorithm~\ref{#1}}

\def\1{\bm{1}}

\DeclareMathAlphabet{\mathsfit}{\encodingdefault}{\sfdefault}{m}{sl}
\SetMathAlphabet{\mathsfit}{bold}{\encodingdefault}{\sfdefault}{bx}{n}

\usepackage{hyperref}
\usepackage{url}
\usepackage{graphicx}
\usepackage{booktabs}
\usepackage{multirow}
\usepackage{xcolor}
\usepackage{subcaption}
\usepackage{algorithm}
\usepackage{algpseudocode}
\usepackage{enumitem}
\usepackage{placeins}
\setlist{nosep,leftmargin=*}

\newcommand{\method}{\mbox{EvoCUE}}

\def\figref#1{Figure~\ref{#1}}
\def\secref#1{Section~\ref{#1}}
\def\eqref#1{Eq.~(\ref{#1})}
\def\algref#1{Algorithm~\ref{#1}}
\newcommand{\appref}[1]{Appendix~\ref{#1}}
\newcommand{\tabref}[1]{Table~\ref{#1}}

\title{The Right Lesson at the Right Step: Deriving Control Updates for Self-Evolving Agents}

\newcommand{\authname}[1]{\makebox[\dimexpr0.5\textwidth-2\tabcolsep-2pt\relax][l]{#1}}

\author{\authname{Yunhe Su\thanks{Equal contribution.}} \\
Independent Researcher \\
\texttt{Yunhe.Su@outlook.com}
\And
\authname{ZiYi Dong\footnotemark[1]} \\
Sun Yat-sen University \\
\texttt{dongzy6@mail2.sysu.edu.cn}
\AND
\authname{Tong Yu} \\
eunomia-bpf community \\
\texttt{yt.xyxx@gmail.com}
\And
\authname{Weijian Deng} \\
Tsinghua Shenzhen International \\
Graduate School, Tsinghua University \\
\texttt{dengwj16@sz.tsinghua.edu.cn}
\AND
\authname{Hao Li} \\
Nankai University \\
\texttt{2120250784@mail.nankai.edu.cn}
\And
\authname{Bowen Jiang} \\
Department of Computer and \\
Information Science \\
University of Pennsylvania \\
Philadelphia, PA, United States \\
\texttt{bwjiang@engineering.upenn.edu}
\AND
\authname{Pengxu Wei\thanks{Corresponding author.}} \\
Sun Yat-sen University \\
Peng Cheng Laboratory \\
\texttt{weipx3@mail.sysu.edu.cn}
}

\iclrfinalcopy %
\begin{document}

\maketitle
\lhead{Preprint}

\begin{abstract}

Self-evolving agents improve future behavior by reusing past experience, typically as global prompts, memories, or reflections. Yet these mechanisms rarely control where experience takes effect. In long tool-use workflows, the same lesson may correct one decision but distract another, making experience reuse a problem of localized control rather than memory alone. We introduce EvoCUE (Evolution through Control Updates from Evidence), a framework for learning reusable control-program updates from completed agent executions. EvoCUE represents the agent as an explicit state-machine controller, whose nodes perform model or tool calls and whose edges define where control passes next. This makes the workflow editable at precise locations, so each learned update can specify what to add, where it acts, and when it applies. From completed trajectories, EvoCUE uses residual goals and observed execution traces to propose localized instruction or skill edits. Each candidate is evaluated at the point where it would act by resuming the parent and edited controllers from the same checkpoint and comparing their final outcomes. Accepted edits are compiled with applicability rules, confirmed on held-out tasks, and inherited by later executions. We evaluate EvoCUE on long tool-use environments where learned conventions must reach the right execution step. From a minimal AppWorld controller without benchmark-specific onboarding instructions, EvoCUE learns the missing task-completion convention and substantially improves success on Test-Normal and Test-Challenge. On PAST-Bench office workflows, EvoCUE transfers organizational requirements from prior episodes to later tasks, improving task-execution quality. These results show that self-evolving agents should place experience inside the control flow, rather than only store it as text.
\end{abstract}

\section{Introduction}
\label{sec:intro}

Language-model agents perform long-horizon tasks through sequences of model calls, tool executions, and routing decisions, such as operating a user's apps to settle payments, update records, or answer questions. In these settings, failures can recur even when the model and tools are capable of the correct behavior. A human worker may infer an organization's conventions after a few failed attempts, but a fixed-model agent repeats the same procedural error unless past experience is converted into a reusable change in its execution procedure. Self-evolving agents aim to provide this ability by improving future behavior from completed executions while keeping the underlying model fixed~\citep{shinn2023reflexion,zhao2024expel,wang2025awm,zhang2026ace}. We study how agents can learn procedural changes from past tasks and apply them where needed.

\begin{figure}[t]
\centering
\includegraphics[width=\linewidth]{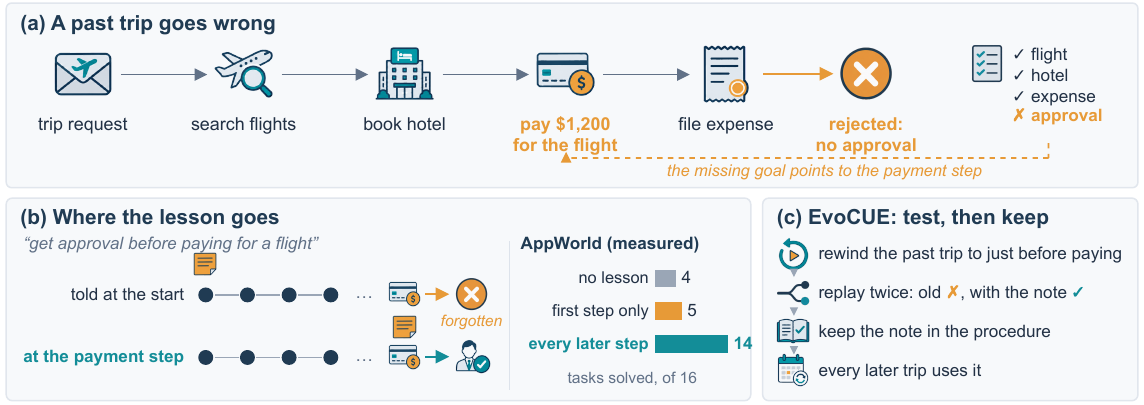}
\caption{Motivation of \method{}. (a) A trip-booking assistant completes the requested operations but pays for the flight without required approval. The missing approval is a residual goal that localizes the failure to the payment step. (b) The lesson helps only when available at the payment decision. The bars show the same effect on AppWorld for a learned instruction about reporting a finished task. (c) \method{} restores the past task to before the edit can act, continues it with and without the edit, and keeps the edit only if it improves the outcome.}
\label{fig:teaser}
\end{figure}

A key difficulty is that useful experience in a long workflow is often local. In the trip-booking example (\figref{fig:teaser}a), the assistant completes the requested operations, but the expense report is rejected because it pays for a flight over \$1,000 without approval. The lesson, ``get approval before paying for a flight,'' helps only if it reaches the payment decision. If shown only at the start of the next trip, it may not affect the later payment step; if attached to that step, it can change the action (\figref{fig:teaser}b). AppWorld~\citep{trivedi2024appworld} shows the same pattern for a real task-completion convention: the same instruction solves 5 of 16 development tasks at the first step, but 14 when available later. Thus, self-evolution must learn not only what to retain, but where it should act.

A second difficulty is measuring whether a proposed change helps. A local edit can influence many later actions, while feedback is usually observed only at the end. Rerunning a task from the beginning can therefore mix the edit's effect with unrelated variation in earlier execution. Existing methods improve agents by optimizing prompts from execution feedback~\citep{zhou2023ape,yang2024opro,khattab2024dspy,opsahlong2024mipro,yuksekgonul2025textgrad,agrawal2025gepa}, searching workflow code or architectures~\citep{hu2025adas,zhang2025aflow,zhang2025maas}, or accumulating memories, playbooks, and skills~\citep{zhao2024expel,wang2025awm,zhang2026ace,suzgun2026cheatsheet,wang2024voyager}. These approaches mainly decide what information or structure to keep. Where that content acts is usually fixed by design, such as a system prompt or retrieval step, and candidates are typically evaluated by full task reruns. This leaves two coupled problems: identifying the control location a lesson should affect, and testing whether an edit at that location improves the final outcome under comparable execution conditions.

We present \method{} (Evolution through Control Updates from Evidence), a framework for learning reusable updates to an agent's control program. \method{} addresses the two challenges above by making both the location and effect of a lesson explicit. It represents the agent procedure as a state machine, whose nodes call the language model or tools and whose edges route execution (\figref{fig:method}). This turns experience into localized controller edits rather than global text: a learned update is an instruction or short executable skill attached to a specific edge, with an applicability rule that determines when it runs. To identify where an edit should act, \method{} estimates residual goals from completed trajectories, namely task requirements that the public execution history does not yet show as satisfied. Residual goals connect failures to editable locations; in the trip example, missing approval points to the payment step. To test whether the edit helps there, \method{} restores a past task to the checkpoint immediately before the edit can act, then continues the task twice, once with the current program and once with the edited program. Because the two executions share the same prefix, their final-score difference estimates the edit's downstream effect under comparable execution conditions; we call this comparison a paired trial (\figref{fig:teaser}c). Edits that pass are compiled with fitted applicability rules, confirmed on separate tasks, and inherited by later tasks.

Starting from a minimal AppWorld controller without the benchmark's hand-written onboarding instructions, \method{} recovers the task-completion convention that official prompts provide manually. Its paired trials reject three plausible alternatives, including a related instruction placed at the first step, before accepting the version that reaches the later completion decision (\figref{fig:trials}). With the model and tools fixed, the learned program raises task success from 13.7\% to 79.2\% on Test-Normal and from 6.7\% to 70.0\% on Test-Challenge. Under the same starting prompt, adapted versions of GEPA, AFlow, AWM, and MaAS learn from all 90 training tasks but improve by at most 4.8 points on Test-Normal. On PAST-Bench office workflows~\citep{xue2026past}, \method{} transfers organizational requirements from prior episodes, such as data-protection approval before export, improving task-execution score from 0.61 to 0.77, while the four baselines change it by at most 0.06. Our contributions are summarized as follows:

\begin{itemize}
\item We introduce \method{}, a framework for learning localized edits to an agent's control program, specifying what to change, where it acts, and when it applies.

\item We propose paired trials from matched checkpoints to test candidate edits at the point where they act, before confirmed edits are inherited by later tasks.

\item We show large gains from a minimal controller on AppWorld and transfer of organizational requirements on PAST-Bench, with the model and tools fixed, and analyze why location matters.
\end{itemize}

\section{Related work}
\label{sec:related}

\textbf{Self-evolving agents and experience reuse.} Reflexion~\citep{shinn2023reflexion} and Self-Refine~\citep{madaan2023selfrefine} turn feedback into verbal self-critique. ExpeL~\citep{zhao2024expel}, CLIN~\citep{majumder2024clin}, AutoManual~\citep{chen2024automanual}, and AutoGuide~\citep{fu2024autoguide} distill insights, rules, or context-conditioned guidelines from past trajectories; Agent Workflow Memory (AWM)~\citep{wang2025awm} induces reusable workflows; ACE~\citep{zhang2026ace} and Dynamic Cheatsheet~\citep{suzgun2026cheatsheet} curate an evolving context; ReasoningBank~\citep{ouyang2026reasoningbank} and Memento~\citep{zhou2025memento} store strategies or cases; Voyager~\citep{wang2024voyager}, Agent Skill Induction~\citep{wang2025asi}, and SkillWeaver~\citep{zheng2025skillweaver} grow libraries of verified skills; LOOP~\citep{chen2025loop} and AgentEvolver~\citep{zhai2025agentevolver} update the model weights with reinforcement learning. These methods mainly decide what to store and deliver it through a channel fixed by the method, typically the prompt or a retrieval step. \method{} keeps the model fixed and treats where content acts as part of the learned edit.

\textbf{Prompt and program optimization.} APE~\citep{zhou2023ape}, OPRO~\citep{yang2024opro}, and Promptbreeder~\citep{fernando2024promptbreeder} search over instructions with model-generated candidates; DSPy~\citep{khattab2024dspy} and MIPROv2~\citep{opsahlong2024mipro} optimize instructions and demonstrations of multi-stage programs; TextGrad~\citep{yuksekgonul2025textgrad} and Trace~\citep{cheng2024trace} propagate textual feedback through computation graphs; GEPA~\citep{agrawal2025gepa} keeps a reflected prompt that improves on its parent over a minibatch. These methods optimize prompts of given modules and compare candidates from the task start; \method{} edits locations within an episode and compares a candidate with its parent from the checkpoint where it acts.

\textbf{Automated design of agents and workflows.} ADAS~\citep{hu2025adas}, AFlow~\citep{zhang2025aflow}, MaAS~\citep{zhang2025maas}, AgentSquare~\citep{shang2025agentsquare}, and GPTSwarm~\citep{zhuge2024gptswarm} search over agent code, workflows, architectures, or computation graphs, and Agent Symbolic Learning~\citep{ou2025symbolic}, G\"odel Agent~\citep{yin2025godel}, and the Darwin G\"odel Machine~\citep{dgm2026} let agents revise their own pipelines or code. StateFlow~\citep{wu2024stateflow} and MetaAgent~\citep{zhang2025metaagent} model agents as finite-state machines, and EvoFSM~\citep{evofsm2026} evolves state-machine structure and state instructions for each query under a language-model critic, retrieving earlier machines from an experience pool. Self-Harness~\citep{zhang2026selfharness} lets an agent edit its own harness and accepts edits after regression tests. \method{} shares the explicit control representation but learns across tasks from scalar outcomes, tests each edit by paired trials from the same checkpoint, fits when the edit applies, and inherits edits only after confirmation.

\paragraph{Interventions and attribution on agent traces.} CANTANTE~\citep{cantante2026} attributes outcomes to individual agents in a fixed workflow using contrastive rollouts. CausalFlow~\citep{bonagiri2026causalflow} and DoVer~\citep{ma2026dover} intervene on steps of failed traces to test repair hypotheses, and ASSAY~\citep{wang2026assay} estimates the effect of each skill by randomized masking and suppresses skills predicted to hurt a task. \method{} instead uses interventions to select persistent control edits: it estimates each edit's downstream effect relative to its parent, fits an applicability rule, and compiles successful edits into the controller for future tasks, turning trace-level evidence into reusable workflow changes.

\section{\method{}: Evolution through Control Updates from Evidence}
\label{sec:method}

\begin{figure}[t]
\centering
\includegraphics[width=\linewidth]{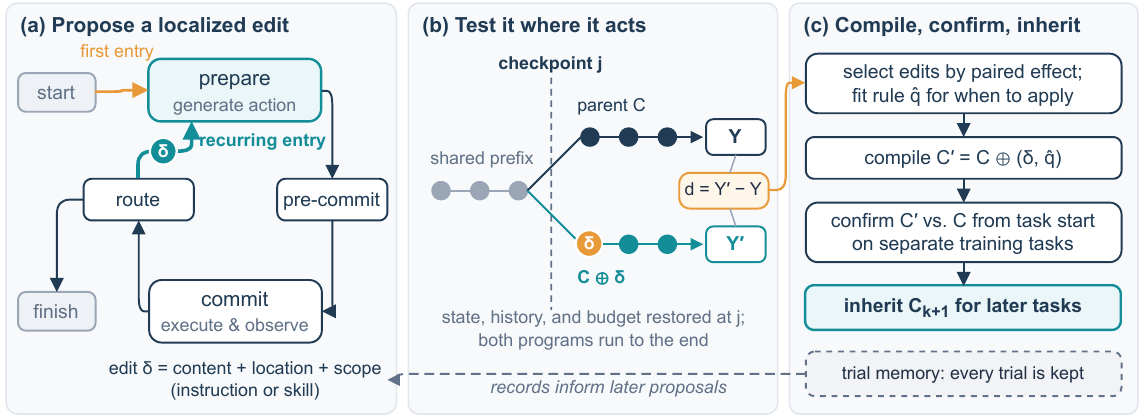}
\caption{\method{} on the AppWorld controller. (a) The controller is a state machine. The action-generation node (prepare) is entered once from start and repeatedly from route; an edit attaches content to one entry, optionally with an applicability rule, or inserts a multi-step skill. (b) A paired trial restores the parent's state at checkpoint $j$, just before the edit's location is reached, and continues the parent and the edited program to the end of the task. (c) Edits that pass are compiled, confirmed on separate tasks, and inherited by later tasks; accepted and rejected trials inform later proposals.}
\label{fig:method}
\end{figure}

\method{} learns a controller through evidence-driven update rounds. In each round, it first uses completed-task experience to propose localized edits to the current control program (\secref{sec:edits}). It then tests each edit where the edit would take effect, using paired trials from matched checkpoints (\secref{sec:paired}). Edits that pass these local tests are compiled into a candidate program, confirmed on separate tasks, and inherited by later executions (\secref{sec:confirm}). \figref{fig:method} illustrates this loop on the AppWorld controller, and \algref{alg:round} summarizes one learning round.

\subsection{Preliminary: Learning control programs from experience}
\label{sec:setup}

\textbf{Controller.} An agent combines a fixed language model and fixed tools with a controller, or control program, $C=(G,\pi,\Omega)$. The graph $G$ has nodes that call the model or a tool and edges, the arrows of the flowchart, that pass control between them; an edge can carry an instruction that the target node receives when it is entered through that edge. The routing policy $\pi$ chooses the next edge from the observable execution state. The library $\Omega$ holds executable skills; like options in reinforcement learning~\citep{sutton1999between}, a skill has a start condition, runs several operations as one step, and returns success or failure. Because edges are explicit, the same node can be entered at different points of an episode. In the AppWorld controller (\figref{fig:method}a), prepare, the action-generation node, writes the next action as code, pre-commit can revise this draft, commit executes it, and route continues or stops. Prepare is entered once from start and again from route after each action; an instruction attached to the first entry is seen once, whereas one attached to the recurring entry is seen at every later step. The edge carrying a lesson thus fixes where it acts and where its test begins.

\textbf{Experience and residual goals.} A completed task yields its public instruction, the actions taken, the observations visible to the agent, and a final scalar score. Public information is what the agent sees while it acts, together with the score revealed after the task ends; hidden tests, reference answers, and grader internals are private and never reach the learner. Residual goals are the requirements of the instruction that the public history does not yet show as satisfied. The language model estimates them from the instruction and the trajectory, marking each requirement as observed or missing with a reference to the supporting observation. For example, if the history shows that a file was exported but contains no approval record, ``obtain approval before export'' remains a residual goal. When every requirement is observed but the score is zero, the failure lies in how the task was completed, as in the AppWorld case of \secref{sec:trajectory}.

\textbf{Initial controller.} Learning starts from an initial controller $C_0$ and never modifies the model, the tools, or the observation interface. In AppWorld, $C_0$ is a six-node controller whose prompt contains general coding guidance and a single sentence about finishing a task: ``Call apis.supervisor.complete\_task with an answer when appropriate only after finishing the requirements.'' It omits the onboarding instructions that the benchmark authors provide with their reference agents, including the convention that tasks without a requested value should not pass an answer. We use it to test whether an agent can acquire such conventions from experience. Gains are measured against the starting controller, and learned programs are frozen before evaluation.

\subsection{Proposing localized edits}
\label{sec:edits}

The learner starts each round with trajectories from completed training tasks
under the current controller $C_k$, along with residual goals, final scores,
restorable checkpoints, and records of earlier trials. A compact index summarizes
these tasks, their outcomes, their residual goals, and prior edit results. The
proposer, implemented as a language model, uses this index and may retrieve full
public fragments of listed trajectories when writing candidate edits. Tasks cited
by a proposal guide which checkpoints are sampled for trials, but task identifiers
never become runtime routing conditions.

A candidate edit $\delta$ specifies its content, its location (the edge where it acts), its scope (how long it stays in effect), and the effect it is expected to have. An instruction edit adds text that a node receives when it is entered through a given edge; its scope is the current call, a skill invocation, or the next action. A structural edit inserts an executable skill of one to six primitive operations, with forward branches and declared success or failure returns, at an edge of $G$. Instructions and skills change $G$ and $\Omega$, and applicability rules (\secref{sec:paired}) change $\pi$. Before any run, we check it uses existing nodes, respects the pending action's state, and gives operators valid inputs (\appref{app:grammar}).

Location and scope determine whether content can act at all. The action-generation node writes an action draft, code that has not yet been executed, and the commit node executes it. A diagnostic operator that only appends a reflection cannot revise a draft that has already been generated, and an instruction attached to the first entry of a node does not reach later entries through another edge. Before execution, the proposer therefore receives feedback that states what each proposed operator reads and writes and which entries an attachment reaches. It describes the controller, not the tasks.

\subsection{Testing edits where they act}
\label{sec:paired}

For each candidate, checkpoints are sampled from parent trajectories of the tasks cited by the proposal and of related failures, before any candidate outcome is observed. A paired trial restores the environment state, public history, controller state, and remaining budget at a checkpoint $j$ just before the edit's location is reached, and then executes both the parent program $C$, which lacks the edit, and the edited program $C\oplus\delta$ to the end of the task. For task $t$ and checkpoint $j$, the observed effect of the edit is
\begin{equation}
 d_{tj}(\delta;C)=Y_{tj}(C\oplus\delta)-Y_{tj}(C),
 \label{eq:effect}
\end{equation}
where $Y$ is the final task score. An edit may act repeatedly and influence the rest of the workflow; $d_{tj}$ measures this total downstream effect. Because both runs share the prefix, the comparison excludes variation in everything that happened before the edit could act.

Each trial record stores its type, both scores, features of the state before the edit acts, whether the edit was reached and applied, and the observed action changes. This execution status distinguishes an unreached location, a skipped edit, a failed attempt, an applied edit, and an unresolved execution. A task failure keeps its score, whereas an unavailable score is treated as missing evidence. Trials that resume from a checkpoint are matched-prefix trials; otherwise, a task-start trial runs both programs from the start, is recorded separately, and carries no checkpoint features.

\textbf{Applicability rules.} For the trials $\mathcal S_\delta$ of an edit, we enumerate rules $q$ made of at most two equality tests on categorical features of the state before the edit acts, such as the execution phase, draft validity, residual-goal status, progress, repetition, and visible completion evidence. Effects are averaged over checkpoints within a task and then over tasks. We select
\begin{equation}
 \hat q_\delta\in\arg\max_{q\in\mathcal Q}\;\hat p_{\mathcal S_\delta}(q)\,\overline d_{\delta,q}-\lambda|q|,
 \label{eq:condition}
\end{equation}
where $\hat p_{\mathcal S_\delta}(q)$ is the fraction of sampled tasks covered by $q$ (tasks with checkpoint states that satisfy $q$), $\overline d_{\delta,q}$ is their mean effect, and $|q|$ is the number of equality tests in $q$; ties prefer shorter rules. The edit passes if at least two tasks are covered, $\overline d_{\delta,\hat q}>0$, and $\hat p_{\mathcal S_\delta}(\hat q)\,\overline d_{\delta,\hat q}-\lambda(|\hat q|+|\delta|)>0$, where $|\delta|$ counts the edit's operations. The empty rule applies the edit wherever its location is reached; task-start trials can only justify the empty rule. This follows the idea of estimating how an intervention's effect varies with the state before the intervention~\citep{athey2016recursive}. With few trials, these estimates are optimistic for the selected rule, so the confirmation step below uses separate tasks.

\subsection{Compiling, confirming, and inheriting}
\label{sec:confirm}

The compiler turns a passing instruction and its rule into a call at the specified edge that runs only when the rule holds, and a passing skill into internal nodes, a start condition, and explicit returns. A later revision at the same location replaces the earlier one; combining two instructions requires proposing the combined content as a new edit. Skill steps consume the normal execution budget.

Passing edits are compiled together into a candidate update, the program $C'$, which is compared with $C$ from task start on a confirmation batch disjoint from the round's proposal and trial tasks. A preconfigured acceptance criterion, applied to the paired score differences on these tasks, decides whether $C'$ is inherited. Confirmation measures the combined program, including interactions among edits and with existing behavior.

\textbf{Trial memory.} Every trial record stores the parent version, the edit, its location and scope, the motivating evidence, the measured effects, and the execution status. Accepted and rejected trials both remain available to later proposals, as compact summaries in the index and as full records on request. The memory lets later proposals revise the content or the location of an earlier idea and avoid directions that have already failed.

\begin{algorithm}[t]
\caption{One \method{} learning round}
\label{alg:round}
\begin{algorithmic}[1]
\small
\Require controller $C$; completed-task experience $H$; trial memory $M$; acceptance criterion
\State Choose tasks for proposals and trials, and a disjoint confirmation set $\mathcal V$
\State Index public trajectories, residual goals, and prior trials from $(H,M)$
\State Propose edits with content, location, scope, and motivating evidence
\State $\mathcal A\gets\emptyset$
\For{each candidate $\delta$}
  \State Sample checkpoints; run paired trials of $C$ and $C\oplus\delta$ (\eqref{eq:effect})
  \State Store every trial record in $M$
  \State Fit an applicability rule $\hat q_\delta$ (\eqref{eq:condition}); \textbf{if} it passes, $\mathcal A\gets\mathcal A\cup\{(\delta,\hat q_\delta)\}$
\EndFor
\If{$\mathcal A\neq\emptyset$}
  \State $C'\gets\operatorname{Compile}(C,\mathcal A)$; compare $C'$ with $C$ from task start on $\mathcal V$
  \State \textbf{if} the acceptance criterion holds, $C\gets C'$
\EndIf
\State Store the confirmation record in $M$; \Return $C,M$
\end{algorithmic}
\end{algorithm}

\section{Experiments}
\label{sec:experiments}

We evaluate \method{} in two settings that test complementary forms of experience reuse. AppWorld tests whether an agent can recover a missing workflow convention from failures in long tool-use tasks, and whether the learned lesson must be placed at the right control location (Sections~\ref{sec:main}--\ref{sec:placement}). PAST-Bench tests whether requirements revealed in earlier office episodes can be carried forward to improve later tasks (\secref{sec:past}).

\subsection{Setup}
\label{sec:exp_setup}

\textbf{AppWorld.} AppWorld~\citep{trivedi2024appworld} contains multi-step tasks over nine simulated apps, solved by writing Python code against 457 APIs. It provides 90 training and 57 development tasks, followed by 168 Test-Normal and 417 Test-Challenge tasks. The official checker decides task success; a scenario succeeds only if all of its tasks succeed. The learner uses a 32-task training subset. Learning uses only training tasks, and paired trials restore checkpoints only on training tasks. The frozen program is evaluated once on Test-Normal and Test-Challenge, which we report in aggregate; no test-set information is used for learning, program selection, or configuration choice.

\textbf{Model and budget.} All model calls use \texttt{deepseek-v4-flash} through a self-hosted endpoint, with temperature 0 (\appref{app:protocol}). Each task allows 40 environment actions and 400 controller steps. The initial and learned controllers share the model, tools, observation interface, termination rules, and budgets; learned instructions and skills consume the same budget.

\textbf{Learning configuration.} A round proposes up to two candidates; each is tested on eight paired tasks, and the compiled update is confirmed on eight separate training tasks. The update is inherited if its mean paired gain on the confirmation tasks is positive. The learner starts from $C_0$ with an empty learned program. It can read public trajectories of earlier training runs under the same $C_0$, including eight model-generated hypotheses from those runs, which enter as untested ideas.

\textbf{Baselines.} GEPA~\citep{agrawal2025gepa}, AFlow~\citep{zhang2025aflow}, AWM~\citep{wang2025awm}, and MaAS~\citep{zhang2025maas} use the same model, tools, observation interface, and budgets; each keeps its own initial program on the same minimal agent prompt as $C_0$ and, on AppWorld, learns from all 90 training tasks. They are not reproductions of the published systems, which use the official prompts (\appref{app:baselines}).

\subsection{One learned instruction improves hundreds of later tasks}
\label{sec:main}

\begin{table}[t]
\caption{Frozen AppWorld task success (\%). \method{} gains 65.5 and 63.3 points. All methods start from the same minimal agent prompt, without official onboarding instructions. Tokens: known learning tokens (\appref{app:baselines}). $^\ast$Final program equals the initial one; differences come from reruns.}
\label{tab:appworld_main}
\centering\small\setlength{\tabcolsep}{2.5pt}
\begin{tabular}{lrrcccc}
\toprule
& & & \multicolumn{2}{c}{Test-Normal ($n=168$)} & \multicolumn{2}{c}{Test-Challenge ($n=417$)} \\
\cmidrule(lr){4-5}\cmidrule(lr){6-7}
Method & Train tasks & Tokens (M) & Initial & Learned & Initial & Learned \\
\midrule
GEPA$^\ast$~\citep{agrawal2025gepa} & 90 & 20.1 & 13.1 & 13.7 & 6.0 & 5.0 \\
AFlow$^\ast$~\citep{zhang2025aflow} & 90 & 112.0 & 8.9 & 13.7 & 5.8 & 4.8 \\
AWM~\citep{wang2025awm} & 90 & 14.2 & 11.9 & 11.9 & 7.2 & 6.0 \\
MaAS~\citep{zhang2025maas} & 90 & 25.0 & 7.7 & 8.3 & 4.3 & 3.1 \\
\midrule
\method{} (ours) & 32 & 15.9 & 13.7 & \textbf{79.2} & 6.7 & \textbf{70.0} \\
\bottomrule
\end{tabular}
\end{table}

\tabref{tab:appworld_main} reports the frozen evaluation. Task success rises from 13.7\% to 79.2\% on Test-Normal, where 110 tasks improve and none regress, and from 6.7\% to 70.0\% on Test-Challenge, with 270 improvements and six regressions. Scenario success rises from 5.4\% to 57.1\% and from 2.2\% to 44.6\%. On the full development set, success rises from 13 to 47 of 57 tasks.

Under this protocol the baselines gain at most eight Test-Normal tasks and none gains on Test-Challenge, despite almost triple the training tasks and 0.9--7.1 times \method{}'s learning tokens. GEPA and AFlow stayed unchanged; AWM's 85 memories and MaAS's controller did not help.

The inherited update is a single instruction attached to the recurring entry of the action-generation node, with an empty applicability rule and no new skill:
\begin{quote}\small
Before calling complete\_task, determine whether the task is a question (asks for a specific value) or an action. If it is a question, provide the requested value as the answer. If it is an action, do not provide an answer (leave it as None).
\end{quote}
It encodes the task-completion convention that the official AppWorld prompts state by hand and that $C_0$ lacks. The edit is reached in 165 of 168 Test-Normal and 413 of 417 Test-Challenge tasks.

\subsection{How the learner found the edit}
\label{sec:trajectory}

\begin{figure}[t]
\centering
\includegraphics[width=\linewidth]{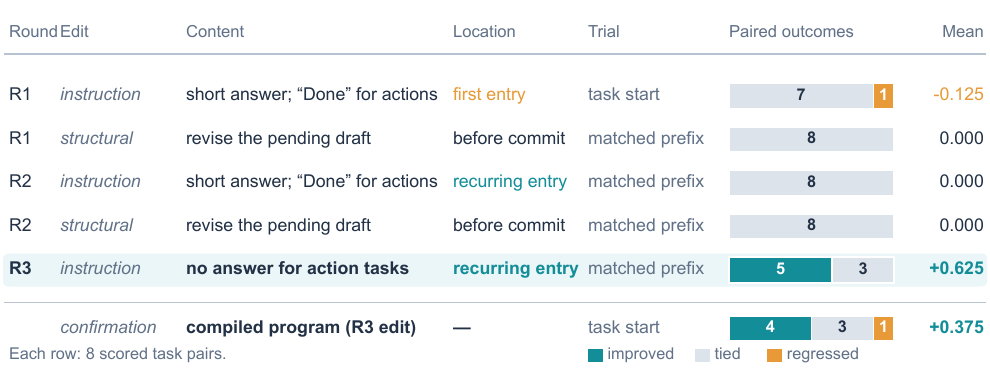}
\caption{Every candidate tested while learning the AppWorld program, with its content, location, trial type, and outcomes on eight scored pairs; the last row is the confirmation on eight separate tasks. Only the completion convention at the recurring entry improves outcomes.}
\label{fig:trials}
\end{figure}

\figref{fig:trials} lists every candidate the learner tested. In the first two rounds, the proposer attributed failures to verbose answers. It asked action tasks to answer with a short confirmation such as ``Done'', first at the first entry, where the edit caused one regression and no improvement, and then at the recurring entry, where all eight pairs tied. A structural edit that inserted a draft-revision step before execution also tied twice. Trial memory recorded these outcomes for later rounds.

In the third round, the residual-goal estimates of failed training tasks provided the decisive evidence. In one of them the agent had commented on and liked every requested payment, and every requirement was marked as observed, yet the task scored zero; in the same code block, the agent had printed the documentation of the completion API and then submitted a one-sentence summary as its answer. The proposer concluded that the actions were done and the failure lay in completion, quoted the documented convention that non-question tasks must leave the answer at its default of None, and attached the resulting instruction to the recurring entry. Five of eight matched-prefix pairs improved and none regressed. All covered tasks improved or tied, so the fitted rule was empty, and the compiled program gained on four confirmation tasks, lost on one, and tied on three. The learner inherited it. Residual goals located the failing step, paired trials separated effective content and location from plausible alternatives, and confirmation decided what entered the program.

\textbf{Rediscovery from scratch.} Two further runs from $C_0$ without the earlier hypotheses, whose proposer may attach one instruction to several entries, rediscover the same convention at both entries of the action-generation node. The first raises Test-Normal success from 11.9\% to 76.2\% (110 tasks improve, two regress); the second solves 48 of 57 development tasks (\appref{app:results}).

\subsection{Where learned content must act}
\label{sec:placement}

We hold the learned content fixed and vary only the steps at which it is available (\figref{fig:teaser}b). The initial controller solves 4 of 16 development tasks. Attached to the first entry of the action-generation node, the instruction solves 5; attached to the recurring entry, as learned, it solves 14. Adding the same text to the agent's prompt at every step also solves 14, and a structured rewrite of the text solves 13. What matters is whether the instruction is available when the completion decision is made, which in a long episode happens many steps after the first action. Eight of the ten newly solved tasks now leave the answer unset instead of submitting a summary (\appref{app:cases}).

Like state-machine and workflow controllers in general~\citep{wu2024stateflow,zhang2025metaagent,zhang2025aflow}, our controller assembles a fresh prompt for each node call, so a first-entry instruction never reaches the completion decision; the explicit controller lets the learner express and test such locations.

\subsection{Transferring organizational requirements to later office tasks}
\label{sec:past}

PAST-Bench~\citep{xue2026past} evaluates personal agents on sequences of simulated office tasks, such as ticket handling and data-sharing requests, whose requirements are revealed through earlier episodes. We adapt it to the same model backend and keep the learned controller as the persistent artifact across episodes. The learner collects 16 past episodes and writes eight knowledge cards, short summaries of requirements that each cite their source episode, as proposal evidence. The first compiled update is rejected after a negative confirmation gain; the second attaches learned instructions to both entries of the action-generation node, with empty applicability rules.

\begin{table}[t]
\caption{PAST-Bench task-execution scores in $[0,1]$. Pilot families: 16 later tasks after learning from 16 past episodes. Unseen families: 8 later tasks from four families never used during development. $^\ast$No update was accepted, so the program equals the initial one. $^\dagger$First of two learning passes.}
\label{tab:past}
\centering\small\setlength{\tabcolsep}{16pt}
\begin{tabular}{lcccc}
\toprule
& \multicolumn{2}{c}{Pilot families} & \multicolumn{2}{c}{Unseen families} \\
\cmidrule(lr){2-3}\cmidrule(lr){4-5}
Method & Initial & Learned & Initial & Learned \\
\midrule
GEPA~\citep{agrawal2025gepa} & 0.599 & 0.628 & 0.279 & 0.300$^\ast$ \\
AFlow~\citep{zhang2025aflow} & 0.594 & 0.582 & 0.308 & 0.292 \\
AWM~\citep{wang2025awm} & 0.590 & 0.564 & 0.303 & 0.441$^\dagger$ \\
MaAS~\citep{zhang2025maas} & 0.557 & 0.620 & 0.316 & 0.327$^\dagger$ \\
\midrule
\method{} (ours) & 0.608 & \textbf{0.772} & 0.305 & 0.332$^\ast$ \\
\method{}, all distilled knowledge & 0.624 & 0.766 & 0.345 & \textbf{0.729} \\
\method{}, learned text at every step & 0.608 & 0.767 & 0.345 & 0.725 \\
\bottomrule
\end{tabular}
\end{table}

On 16 later pilot tasks, the frozen program raises the task-execution score from 0.608 to 0.772, a gain 2.6 times the largest baseline gain, with 11 improvements, one regression, and four ties (\tabref{tab:past}). In a data-sharing request, the learned agent tags the ticket as pending, assigns the data-export category, and records that approval from the data protection officer is required before export, which raises the task's score from 0.68 to 0.85. The same learned text given at every step obtains 0.767, again showing that the learned content, once it reaches the relevant steps, carries the improvement. Two repeated evaluations of both programs give 0.770 and 0.773 (initial: 0.625 and 0.621).

\textbf{Unseen task families.} On four task families never used during development and fully supported by our adapter (8 later tasks), the knowledge cards that \method{} distills from their past episodes, installed at the learned entries, raise the task-execution score from 0.35 to 0.73, for example from 0.10 to 0.84 on an incident-handling family, whereas the four baselines gain at most 0.14 (\tabref{tab:past}, \appref{app:results}). With only two improvable confirmation tasks, paired confirmation accepts none of these edits; with little confirmation data, it trades recall for protection against regressions.

\subsection{Six further domains}
\label{sec:boundaries}

On $\tau^2$-bench~\citep{barres2025tau2}, EvoClawBench~\citep{evoclawbench2026}, MBPP~\citep{austin2021mbpp}, HumanEval~\citep{chen2021humaneval}, MATH~\citep{hendrycks2021math}, and HotpotQA~\citep{yang2018hotpotqa}, the initial programs already score high or the tasks are short (\tabref{tab:domains}, \appref{app:results}). \method{} accepts no update, so its controller remains $C_0$; the baselines also change little; apart from EvoClawBench runs with execution errors, the largest change is GEPA's $+3.1$ points on MATH.

\section{Discussion and conclusion}
\label{sec:conclusion}

We present \method{}, a framework for learning reusable control-program updates from completed agent executions. By representing the agent as an explicit state machine, \method{} makes each learned edit specify what to change, where it acts, and when it applies. Residual goals point to the step where the current procedure failed, and paired trials from matched checkpoints, in which the parent and edited programs share the execution prefix before the edit can act, measure each edit's effect where it acts. These trials select useful edits and fit applicability rules; confirmation on separate tasks decides which updates later tasks inherit.
From a minimal AppWorld controller, \method{} learns the task-completion convention that benchmark prompts state by hand, raising Test-Normal success from 13.7\% to 79.2\% with the model and tools fixed; the same instruction solves 14 of 16 development tasks when available at every later step but only 5 at the first step. On PAST-Bench, it transfers organizational requirements from prior episodes to later office tasks, improving execution quality beyond adapted baselines. This shows that experience-driven adaptation should learn not only what experience to reuse, but also where it should enter the agent's control flow.

\textit{Limitations and future directions.}
Our work leaves several questions open. Learned applicability rules are inferred from a finite set of observed tasks. Although our results demonstrate transfer to subsequent tasks, broader transfer and systematic handling of exceptions remain open problems. Future work could learn explicit representations of the contexts in which requirements hold, fail, or require modification. Richer structural edits, conditional applicability rules, and interactions among multiple updates are also promising directions for future research. Furthermore, paired trials require restorable execution states, which may not be available in all environments. Our current evaluation also freezes the learned programs after training; a more challenging setting is online learning over long streams of tasks, where lessons accumulate over time and repeatedly injecting an ever-growing set of instructions at every step becomes increasingly costly and unwieldy. We expect explicit locations to become particularly important in this setting, as they allow learned experience to be applied selectively rather than repeatedly exposed to the agent throughout every task.

\clearpage

\subsection*{AI use statement}
Generative AI tools assisted with implementation and manuscript polishing. Language models are also components of the evaluated agents and learners, as specified in \appref{app:protocol}. The icons in Figure~\ref{fig:teaser} were generated with an image-generation model. Figure~\ref{fig:teaser}a,c shows an illustrative example; all other numbers, program structures, and trajectory content shown in the figures come from recorded runs. We take responsibility for the final content of this work, including text, claims or artifacts produced with the aid of generative AI.

\subsection*{Ethics statement}
The experiments use benchmark software environments and simulated office tasks. Private evaluation information is separated from the learner's public interface, and training scores become available only after a task is completed. Learned updates can cause regressions; the restricted edit grammar, recorded interventions, and whole-program confirmation support inspection of this behavior but do not establish deployment safety outside the evaluated settings.

\subsection*{Reproducibility statement}
\appref{app:protocol} specifies the data roles, model configuration, learning settings, selection rules, and implementation identities. \appref{app:grammar} describes the edit grammar and execution semantics, and \appref{app:learned} records the learned programs and their provenance. Additional results, resource use, and execution cases appear in \appref{app:results} and \appref{app:cases}. Every reported number is tied to a recorded configuration and data role, and frozen evaluation is kept separate from learning.

\bibliography{references}
\bibliographystyle{iclr2027_conference}

\newpage
\appendix
\section{Experimental details}
\label{app:protocol}

\subsection{Data roles and public observations}
AppWorld~\citep{trivedi2024appworld} uses its official training, development, Test-Normal, and Test-Challenge splits. The reported learning run uses 32 training tasks, which extend an earlier 16-task diagnostic subset with 16 tasks, and a 16-task development subset for configuration choice. The full development set contains 57 tasks. Test-Normal and Test-Challenge contain 168 and 417 tasks, respectively, and are evaluated with the frozen program. The official checker determines task success; a scenario succeeds only if all its task instances do.

The agent receives the public task instruction, supervisor information, and the printed output of generated Python code, including API documentation it explicitly queries and prints. Unprinted API return values are not automatically exposed to the agent or the learner. Final scalar scores are available after a training task finishes. Private checker details, reference answers, and hidden tests are excluded from learning, program selection, and configuration choice.

PAST-Bench~\citep{xue2026past} is adapted to the same model backend. The learned controller is the persistent artifact across tasks; this differs from the benchmark's native setting, which persists the agent's home directory across tasks. The pilot collects 16 historical episodes, uses eight later episodes as training tasks for learning, and evaluates the frozen program on a separate 16-task pilot previously inspected during development. We report the benchmark's task-execution score, scored with the benchmark rubrics and our backend model as the language-model judge instead of the benchmark's default judge model. The composite score, which also includes persistence components, rises from 0.402 to 0.598 on the frozen pilot. These eight training tasks belong to the learning pool and are not a generalization test.

\subsection{Model and execution settings}
All model calls use \texttt{deepseek-v4-flash} through the same self-hosted endpoint. Decoding uses low reasoning effort, a 4,096-token thinking budget, an output limit of 32,768 tokens, temperature 0, and top-$p$ 1. AppWorld permits 40 environment actions and 400 controller steps; the PAST pilot permits 900 controller steps. The AppWorld initial and learned controllers use the same model configuration, observation protocol, and budgets.

\begin{table}[h]
\caption{Settings of the reported learning instances. Historical hypotheses are prior model-generated training proposals, not instructions added to the initial controller.}
\label{tab:hparams}
\centering\small
\begin{tabular}{lrr}
\toprule
Setting & AppWorld & PAST pilot \\
\midrule
Experience used for this instance & 32-task pool & 16 episodes \\
Maximum learning rounds & 3 & 2 \\
Maximum candidates per round & 2 & 2 \\
Paired tasks per candidate & 8 & 4 \\
Whole-program confirmation tasks & 8 & 4 \\
Prior training hypotheses & 8 & 0 \\
Maximum equality tests per rule & 2 & 2 \\
\bottomrule
\end{tabular}
\end{table}

\subsection{The reported AppWorld run}
The learning run starts from an empty learned program but can read earlier training experience under the same initial controller: public trajectories and eight hypotheses taken from five model-generated proposal outputs of earlier training runs. 

The AppWorld acceptance criterion is a positive mean paired task-score difference on the confirmation tasks. PAST uses a stricter criterion: positive mean gain, at least two improved tasks, and non-negative total gain after removing the largest improvement. Both criteria select updates during training. Within a round, confirmation tasks are disjoint from the tasks used for proposals and trials; tasks reused across rounds remain training data.

\textbf{Trials used for rule fitting.}
In AppWorld, rule fitting uses trials in which the edit executed and a behavior change was recorded, together with scored failed attempts. For the accepted candidate, all eight matched-prefix trials applied the edit, but one successful tied pair showed no observed action change and was excluded by this filter. The fitted empty rule therefore rests on seven tasks with mean difference $5/7$, whereas the figures report all eight scored pairs, with mean difference $5/8$. The complete trial record retains both the inclusion decision and the observed difference.

\textbf{Feedback available to proposals.}
Trial memory retains the full trial effects, the parent version, the edit, and the execution status. The AppWorld proposer receives compact summaries of accepted and rejected trials in the index and can request full records. In rounds two and three it did not request the full pairwise comparisons, so its later proposals used the summaries, while measured effects entered the pass decision, rule fitting, and confirmation directly. Giving every proposal the full quantitative records is a different configuration that these results do not evaluate.

\subsection{Implementation identity}
The AppWorld launch commit is \texttt{e92802f}, with its saved source snapshot. Its initial protocol is traced to \texttt{805ab65}. The learned artifact has SHA-256 prefix \texttt{fe242766c4b7374b}; the launch source manifest has prefix \texttt{f665a2c03271600e}; the learning configuration has prefix \texttt{70bc44a2adb053cb3}. Full hashes and source paths accompany the result ledger. The PAST pilot uses run \texttt{459443c}, learner \texttt{3c6b34f}, and benchmark \texttt{f8223517}.

\section{Edit grammar and execution semantics}
\label{app:grammar}

\subsection{Primitive operations}
A draft is unexecuted action text. Its state is empty, pending-valid, or pending-invalid.
\begin{center}\small
\begin{tabular}{p{2.3cm}p{2.2cm}p{6.9cm}}
\toprule
Operator & Draft effect & Execution meaning \\
\midrule
Prepare & stores draft & Generate an action and store it. \\
Test & preserves draft & Check syntax without an extra model call. \\
RefinePending & revises draft & Rewrite the stored draft; retain it on parse failure. \\
Commit & consumes draft & Execute the action and append its public observation. \\
Reflect & preserves draft & Append a diagnosis to public history. \\
Step variants & generate and execute & Invoke the original combined generation and execution operation, optionally with selection or refinement. \\
\bottomrule
\end{tabular}
\end{center}
Prepare and Commit split the original combined step into two operations so that edits can act between generation and execution. Without a learned edit, this adds no generation call, default check, or retry. Type checks constrain execution semantics without task-specific repairs.

\subsection{Candidate forms}
An instruction candidate specifies a source node and a target node, which together define an edge, the instruction text, and its scope. Scope may be the current call, a skill invocation, or the next action. A structural candidate specifies an entry location, one to six operations, forward conditional branches, a return node, and failure returns for reachable draft states. Each candidate can cite training tasks and state a hypothesis. Cited tasks guide trial sampling but never become routing conditions.

The compiler rejects nonexistent nodes, incompatible draft states, invalid branch indices, and undeclared returns. A later operation can consume valid outputs of earlier steps. A reflection that only writes history is distinguished from a revision that changes the pending draft. These semantics let the learner express a proposed edit faithfully.

\subsection{Effect records and rules}
A trial record includes parent and candidate identifiers, task, trial type (matched-prefix or task-start), features of the state before the edit acts, outcomes, execution status, and observed behavior changes. Missing scores are distinct from zero scores. A task-start trial carries no checkpoint features. Matched-prefix and task-start trials are kept separate.

Rules use categorical features of the state before the edit acts, such as execution phase, draft validity, residual-goal status, progress, repetition, and visible completion evidence. The AppWorld feature set includes detection of a completion API call. Neither task IDs nor private scores can enter runtime conditions. A rule contains at most two equality tests on typed features. Task-wise averaging and coverage implement \eqref{eq:condition}, with $\lambda=10^{-3}$. The term $\lambda|\delta|$ appears only in the passing condition, so $|\delta|$ does not affect which rule is selected. Task-start evidence can only justify the empty rule.

\subsection{Overlapping updates and skill returns}
An instruction revision attaches to a source--target edge. The latest applicable revision supplies the instruction and scope at that edge. A combined instruction must be proposed and tested as a new candidate. Skills add internal nodes and edges, an entry with its applicability rule, a start condition, returns, and a step limit. More specific internal branches take precedence; an explicit unconditional branch precedes the sequential default. Failure returns preserve the declared draft compatibility.

Top-level routing applies learned updates in order. Gain-based selection within an update uses evidence relative to that update's parent. Different candidates can have different sampling pools, so this is a selection heuristic; gains from different parents are not treated as directly comparable. Every top-level and skill operation consumes the effective execution budget.

\section{Learned programs and their provenance}
\label{app:learned}

\subsection{AppWorld: a recurring completion instruction}
The selected program retains six nodes and six edges and adds one instruction at the recurring \texttt{route}$\rightarrow$\texttt{prepare} entry, with call scope and an empty applicability rule. The first action is generated through the original first entry. Later action-generation calls receive the learned instruction:
\begin{quote}\small
Before calling complete\_task, determine whether the task is a question (asks for a specific value) or an action. If it is a question, provide the requested value as the answer. If it is an action, do not provide an answer (leave it as None).
\end{quote}
The question/action distinction is expressed in learned instruction content, not a hand-written runtime classifier or a fitted non-empty rule. The program adds no callable skill.

The learner's public evidence contains training code that explicitly prints the documentation of the task-completion API, which states that the answer ``must be left to the default value, i.e., None'' when the task is not a question. In the third round, the proposer cites this documentation together with residual-goal estimates showing that every requirement of failed training tasks had been observed. Earlier model proposals included both an empty-answer hypothesis and the conflicting proposal to return a completion message for action tasks. Eight historical hypotheses, derived from five actual proposal outputs, entered learning as hypotheses to test. They were not added to the initial controller.

The successful candidate is generated in the third round and keeps its recurring entry after the feedback on which entries an attachment reaches (\secref{sec:edits}). Eight matched-prefix trials give five improvements and three ties. Rule fitting selects the empty rule; a separate eight-task task-start confirmation gives four improvements, one regression, and three ties. The accepted program is then frozen. This chain connects the proposed content, its location, the trial evidence, and later reuse.

\subsection{PAST: organizational requirements at two entries}
The PAST instance collects 16 public historical episodes and writes eight knowledge cards that cite their source episodes. These cards enter the proposal evidence. The first proposed update is rejected after a mean confirmation difference of $-0.030$; the second gives mean paired improvement $0.151$ and confirmation improvement $0.144$. Its instructions are attached to both the first and the recurring entries of the action-generation node. The applicability rules are empty and no new skill is added.

These artifacts illustrate two kinds of learned content: a cross-task execution convention in AppWorld and requirements derived from historical office tasks in PAST. Both become instructions for later tool execution. They do not require graph growth to constitute a learned program update.

\section{Baselines}
\label{app:baselines}

\textbf{Adaptation.} GEPA~\citep{agrawal2025gepa}, AFlow~\citep{zhang2025aflow}, AWM~\citep{wang2025awm}, and MaAS~\citep{zhang2025maas} run on the same model endpoint, tools, observation interface, and per-task budgets as \method{}, and receive the same public information: task instructions, the agent's observations, and scores after a training task ends. Each keeps its own initial program on the same minimal agent prompt as $C_0$, which does not contain the benchmark's official completion instructions, and learns a different object. GEPA revises the agent instruction through reflection on traces and minibatch comparison with its parent. AFlow searches workflow code with Monte Carlo tree search. AWM induces workflow memories from successful trajectories and adds them to the agent prompt. MaAS trains a controller that selects operator compositions for each query. Each method uses its native stopping rule and, where applicable, at most six rounds of four tasks.

\textbf{AppWorld.} All four baselines learn from the 90 training tasks; \method{} uses the 32-task subset described in \secref{sec:exp_setup}. Their learning outcomes differ. GEPA's own validation rejected its reflection proposals, so its final instruction equals its initial (empty) instruction; some rejected proposals placed task-specific text in the general instruction. AFlow searched for six rounds and selected its initial workflow. AWM added 85 workflow memories in one pass over the training tasks, and MaAS trained its controller on 23 batches covering all 90 tasks. Known learning tokens and model calls are 20.1M and 2,116 for GEPA, 112.0M and 10,766 for AFlow, 14.2M and 1,135 for AWM, 25.0M and 3,361 for MaAS, and 15.9M and 1,300 for \method{}; calls without reported usage are not imputed. Learned-baseline scenario success is at most 7.1\% on Test-Normal and 2.2\% on Test-Challenge.

\textbf{Relation to published results.} These runs measure how well each method learns from the minimal agent prompt. They are not reproductions of the published systems, which start from the benchmark's official prompts, and published numbers with official prompts are higher; for example, DARC~\citep{wang2026darc} reports 54.2\% Test-Normal success for GEPA with the same model. We therefore make no claim about the original methods beyond the protocol stated here.

\textbf{PAST-Bench and further domains.} On PAST-Bench, all methods learn from the same 16 past episodes and are evaluated with their frozen programs on the same 16 later pilot tasks (\tabref{tab:past}). On each further domain (\tabref{tab:domains}), all methods use the same learning pool and test tasks.

\textbf{Learned text versus location.} A comparison with text-based experience learning must let that method discover and revise its own experience from public histories. This differs from the location variants in \figref{fig:teaser}b, which receive an instruction already learned by \method{} and vary only the steps at which it is available.

\section{Additional results and resource use}
\label{app:results}

\subsection{Task and scenario success}
The full development set gives 13 of 57 successes for the initial controller and 47 for the learned program. The development set supports diagnosis and is reported apart from the test sets.

\subsection{Rediscovery without earlier hypotheses}
Two further AppWorld learning runs start from $C_0$ without the eight earlier hypotheses, with a proposer that may attach one instruction to several entries of a node. Both learn the completion convention and attach it to the first and the recurring entry, in their second and third rounds respectively. The first program solves 128 of 168 Test-Normal tasks against 20 for a concurrent evaluation of $C_0$ (110 improved, 2 regressed; mean gain 0.64, 95\% bootstrap interval $[0.54, 0.74]$) and 47 of 57 development tasks; the second solves 48 of 57 development tasks. Three independent seeds of an earlier version of the learner, evaluated under its original protocol, each learn the same convention and solve 98 to 114 of 168 Test-Normal tasks, against 16 to 19 for their initial controllers.

\subsection{PAST-Bench: unseen task families}
The unseen-family study selects ten PAST-Bench families that were not used during development, with the pilot's episode-selection rule (two past and two evaluation episodes per family). Six of them seed each episode with preloaded memories and earlier sessions that our adapter cannot load; on these, every method changes the score by at most 0.02. \tabref{tab:past} reports the four fully supported families (8 evaluation tasks; 7 for AWM, which left one task unfinished). AWM and MaAS completed the first of their two learning passes. Over all ten families, installing all of \method{}'s distilled knowledge raises the score from 0.31 to 0.47.

\subsection{Six further domains}
\begin{table}[h]
\caption{Six further domains: initial$\to$learned score of each method on the test tasks, with the model and tools fixed. $\tau^2$-bench (airline and retail) and EvoClawBench are 16-task pilots. $^\ast$\method{} accepted no update, so its learned controller equals $C_0$ and the difference is repeated execution. $^\dagger$No update accepted; the unchanged controller was not re-evaluated on the test tasks.}
\label{tab:domains}
\centering\footnotesize\setlength{\tabcolsep}{2.5pt}
\begin{tabular}{llccccc}
\toprule
Domain ($n$) & Metric & GEPA & AFlow & AWM & MaAS & \method{} \\
\midrule
$\tau^2$-bench (16) & solved & 10$\to$11 & 14$\to$14 & 12$\to$12 & 11$\to$12 & 13$\to$11$^\ast$ \\
EvoClawBench (16) & score & 1.000$\to$0.992 & 1.000$\to$0.499 & 0.999$\to$0.938 & 0.688$\to$0.938 & 0.998$\to$1.000$^\ast$ \\
MBPP (341) & passed & 324$\to$325 & 323$\to$322 & 319$\to$318 & 313$\to$319 & 323$\to$322$^\ast$ \\
HumanEval (131) & passed & 127$\to$127 & 127$\to$127 & 128$\to$127 & 124$\to$126 & no update$^\dagger$ \\
MATH (486) & solved & 411$\to$426 & 413$\to$408 & 407$\to$408 & 387$\to$391 & no update$^\dagger$ \\
HotpotQA (800) & F1 & 0.785$\to$0.785 & 0.783$\to$0.795 & 0.781$\to$0.798 & 0.781$\to$0.781 & 0.786$\to$0.782$^\ast$ \\
\bottomrule
\end{tabular}
\end{table}

\tabref{tab:domains} reports all five methods on six further domains, each with its own initial program and the same model and tools. \method{} accepts no update in any of them. On MBPP, HotpotQA, $\tau^2$-bench, and EvoClawBench its learned controller equals $C_0$, so its differences reflect repeated execution; on HumanEval and MATH the unchanged controller was not re-evaluated on the test tasks. The $\tau^2$-bench pilot uses eight held-out airline and eight retail tasks, with the user simulated by the same model; simulated dialogues differ between runs, and four evaluations of the same initial controller solved 11 to 13 of the 16 tasks. The EvoClawBench pilot uses 16 automated tasks with the official graders in isolated containers, but our adaptation does not enforce the benchmark's whole-task time limit. Its large baseline changes coincide with execution failures: all five zero scores of MaAS's initial program and all eight zero scores of AFlow's learned program occur in runs with execution errors. Otherwise, the largest baseline change is GEPA's $+3.1$ points on MATH.

\subsection{Measured resources}
\begin{table}[t]
\caption{Recorded inference resources for the initial and learned programs. Tokens are known totals in millions; missing denotes calls without reported token usage. AppWorld rows are the frozen test evaluations, and PAST rows are the 16-task frozen pilot evaluation.}
\label{tab:cost}
\centering\small
\begin{tabular}{llrrr}
\toprule
Evaluation & Program & Calls & Known tokens (M) & Missing \\
\midrule
AppWorld Normal & Initial & 2,869 & 31.56 & 21 \\
 & Learned & 3,214 & 34.98 & 49 \\
AppWorld Challenge & Initial & 8,166 & 117.76 & 78 \\
 & Learned & 8,683 & 125.28 & 86 \\
PAST pilot & Initial & 208 & 1.17 & 0 \\
 & Learned & 146 & 0.56 & 0 \\
\bottomrule
\end{tabular}
\end{table}

The AppWorld learning artifact records 1,300 model calls, 15.88 million known tokens, and 21 calls with unavailable usage. This record does not include all earlier exploration that produced the historical hypotheses, so it is not a complete end-to-end learning cost. The PAST pilot learning record contains 555 calls and 2.57 million known tokens with no missing usage. Resource totals include actual calls with reported usage; missing token counts are not imputed as zero.

\section{Execution and reuse cases}
\label{app:cases}

\subsection{AppWorld development behavior}
On the 16-task development comparison, the initial and learned programs solve four and 14 tasks. Among the ten improvements, eight change an action-task completion from a summary string to an unset answer, one newly invokes the completion API with an unset answer, and one question returns the requested value without additional wording. Another task changes its completion argument but still fails, illustrating that the instruction does not replace the task's actual operations.

The two trajectories can also differ in earlier actions and length. These observations connect the learned instruction to execution behavior, while the paired training trials supply the edit-level outcome comparisons. The complete 16-task runs use 242 and 263 model calls and 117 and 131 environment actions for the initial and learned programs, respectively. Budgets are fixed, but realized inference costs can differ between the two programs.

\subsection{PAST frozen pilot}

The PAST pilot uses a separately learned controller, not the AppWorld artifact. The learned instructions are used 65 times in the 16-task frozen evaluation. Eleven task-execution scores improve, one declines, and four are unchanged. The same learned text given at every step obtains $0.767$, compared with $0.772$ for the learned program; the learned content, once it reaches the relevant steps, carries the improvement (\tabref{tab:past}).

In a simulated data-sharing request, the learned agent applies \texttt{data-share-pending}, assigns medium priority and a data-export category, and records that approval from the data protection officer is awaited before export. The task-execution score changes from $0.676$ to $0.851$. In a shared-outage workflow, the score changes from $0.404$ to $0.864$. A far-transfer task regresses from $0.796$ to $0.510$, so the learned program's reuse is not uniformly beneficial.

\end{document}